# Image Identification Using SIFT Algorithm: Performance Analysis against Different Image Deformations

Ebrahim Karami[1], Mohamed Shehata[1], and Andrew Smith[2]
[1]Faculty of Engineering and Applied Sciences, Memorial University, Canada
[2]Faculty of Medicine, Memorial University, Canada

*Abstract*- Image identification is one of the most challenging tasks in different areas of computer vision. Scale invariant feature transform is an algorithm to detect and describe local features in images to further use them as an image matching criteria. In this paper, the performance of the SIFT matching algorithm against various image distortions such as rotation, scaling, fish eye and motion distortion are evaluated and false and true positive rates for a large number of image pairs are calculated and presented. We also evaluate the distribution of the matched keypoint orientation difference for each image deformation.

## I. INTRODUCTION

Image object classification is an important task in the areas of machine vision and especially in remote sensing and is traditionally performed by extracting a set of the texture and shape features. Scale Invariant Feature Transform (SIFT) is a feature detector developed by Lowe in 2004 and has proven to be very efficient in object recognition applications [1]. Speed up Robust Feature (SURF), and Oriented SIFT and Rotated BRIEF (ORB) are other scale- and rotation-invariant interest point detector and descriptors [2-4]. The SIFT feature extraction has four main steps. First is to estimate a scale space extrema using the Difference of Gaussian (DoG). Secondly, a key point localization where the key point candidates are localized and refined by eliminating the low contrast points. Thirdly, a key point orientation assignment based on local image gradient and lastly a descriptor generator to compute the local image descriptor for each key point based on image gradient magnitude and orientation [1].

Optimal matching of the SIFT descriptors is still an open problem. There are several modified matching technique for the SIFT [5-11]. For obtain optimal matching for the SIFT descriptors, we first need to know the statistics of the matched keypoints such that we can remove outliers from the set of matches and improve the accuracy of matching. In this paper, the matching performance of SIFT descriptors against different image deformations is studied by evaluation of the false positive and true positive rates.

The rest of this report is organized as follow. In Section II, SIFT is briefly introduced. In Section III, the matching performance of SIFT against various image deformations is presented. The report is concluded in Section IV.

## II. SIFT ALGORITHM

Scale Invariant Feature Transform (SIFT) was presented by Lowe [1]. The SIFT algorithm transforms the image into a collection of local feature vectors. These feature vectors are aimed to be distinctive and invariant to any scaling, rotation or translation of the image.

In the first step, the feature locations are determined as the local extrema of Difference of Gaussians (DOG pyramid) as given by (3). To implement the DOG pyramid the input image is convolved iteratively with a Gaussian kernel (2). This procedure is repeated as long as the down-sampling is possible. Each collection of images of the same size is called an octave. All octaves build together the so-called Gaussian pyramid by (1), which is represented by a 3D function $L(x, y, \sigma)$:

$$L(x, y, \sigma) = G(x, y, \sigma) * I(x, y), \quad (1)$$

$$G(x, y, \sigma) = \frac{1}{2\pi\sigma^2} e^{-(x^2+y^2)/2}, \quad (2)$$

$$\begin{aligned} D(x, y, \sigma) &= (G(x, y, \sigma) - G(x, y, \sigma)) * I(x, y) \\ &= L(x, y, k\sigma) - L(x, y, \sigma). \end{aligned} \quad (3)$$

The local extrema (maxima or minima) of DOG function are detected by comparing each pixel with its 26 neighbors in the scale-space as Figure 1. The search for extrema excludes the first and the last image in each octave because they do not have a scale above and a scale below respectively. Scale-space extrema detection produces too many keypoint candidates, where some of which are unstable and less useful. In the next step, a detailed fit is performed to the nearby data to find the accurate location, scale, and ratio of principal curvatures. This information is useful to the points which have low contrast or

For each candidate keypoint, interpolation of the nearby data is used to accurately estimate its position. The interpolation is done using the quadratic Taylor expansion of the Difference-of-Gaussian scale-space function, $D(x, y, \sigma)$ with the candidate keypoint as the origin. This Taylor expansion is given as

$$D(x) = D + \frac{\delta D^T}{\delta x} x + \frac{1}{2} x^T \frac{\delta^2 D}{\delta x^2} x, \quad (4)$$

where D and its derivatives are evaluated at the candidate keypoint and x=(x, y, σ) is the offset from this point.

In the next step, for each keypoint, one or more orientations are assigned based on local image gradient directions. This is a useful step in achieving invariance to rotation as the keypoint descriptor can be represented relative to this orientation and therefore achieves invariance to image rotation. First, the Gaussian-smoothed image $L(x, y, \sigma)$ at the keypoint scale $\sigma$ is taken so that all computations are performed in a scale-invariant manner. For an image sample $L(x, y)$ at scale σ, the gradient magnitude, $m(x, y)$, and orientation, $\theta(x, y)$, are precomputed using pixel differences as

$$m(x,y) = \sqrt{(L(x+1,y)-L(x-1,y))^2 + (L(x,y+1)-L(x,y-1))} \quad (5)$$

$$\theta(x,y) + \tan^{-1}\left(\frac{L(x,y+1)-L(x,y-1)}{L(x+1,y)-L(x-1,y)}\right) \quad (6)$$

## III. PERFORMANCE EVALUATION

In this Section, performance of SIFT algorithm for incorrect and correct matches are evaluated. For each image deformation, true positive rate is computed and the statistics of the orientation different between the matched keypoint are also studied. This will be useful for further work on the optimization of the matching for SIFT algorithm.

### A. Incorrect Match (False Positive)

In order to calculate false positive rate versus matching rate threshold $r_T$, we need to compare keypoints of different image pairs and find the matching rate for each comparison. For this experiment, 40000 image pairs (i.e., $200 \times 200$) from Nister database were considered. Figure 2 present false positive rate $P_F(r_T)$ versus threshold $r_T$. From this figure, one can see when $r_T > 0.2$, false positive rate is very close to zero and consequently, a matching rate larger than 0.2 can guarantee a correct match. Figure 3 presents the distribution of $\delta\varphi$, where $\delta\varphi$ is the difference between orientations of each matched keypoints. From this Figure, one can easily see that when images are incorrectly matched, $\delta\varphi$ is well distributed in the entire ranges and it takes its lowest values at multiple integers of 45 degrees, i.e. each two SIFT histogram of orientation bins.

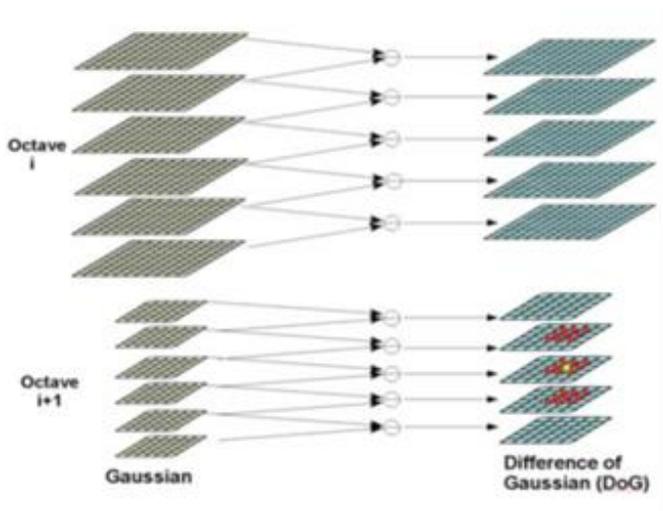

Figure 1. The scale space of SIFT [1].

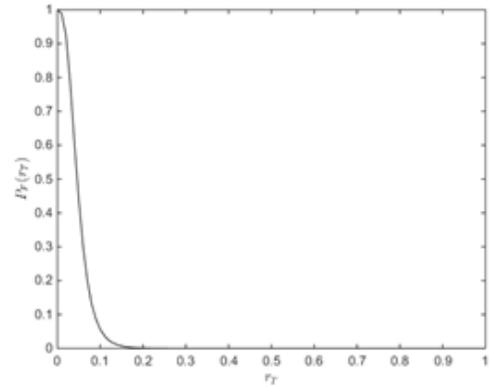

Figure 2. False positive versus matching ratio threshold.

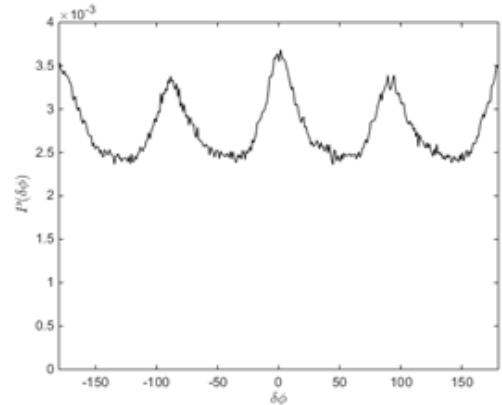

Figure 3. Probability of matched keypoint orientation difference for incorrect matches.

### B. Rotation

Figure 4 presents true positive rate versus matching rate threshold. From this figure, one can easily see that with 30, 60, 90, and 120 degrees rotation, the same performance is achieved. But the best performance among these rotation angles belongs to 90 degrees scenario. This is because SIFT algorithm performs the best at rotations equal to multiple integers of 22.5 degrees which the size of each bin in the histogram of gradient employed by the SIFT algorithm. From figure 5, one can see that the sharpest peak belongs to 90 degrees scenario which confirms the previous result.

### C. Scaling

For this experiment, 1000 images from Nister database are scaled with scaling factors $\alpha =1$ and 2, and true positive rate and $P(\delta\varphi)$ for each scenario are calculated. Figure 6 presents true positive rate versus matching rate threshold. From this figure, one can easily see that a larger scaling factor results a larger matching ratio. This can be due to the larger number of the keypoints in the images with a larger scaling factor which provides a higher chance to find a match between keypoint pairs. From figure 7, one can see that $P(\delta\varphi)$ does not depend on the scaling factor and both scenarios provide the same keypoint orientation difference profile.

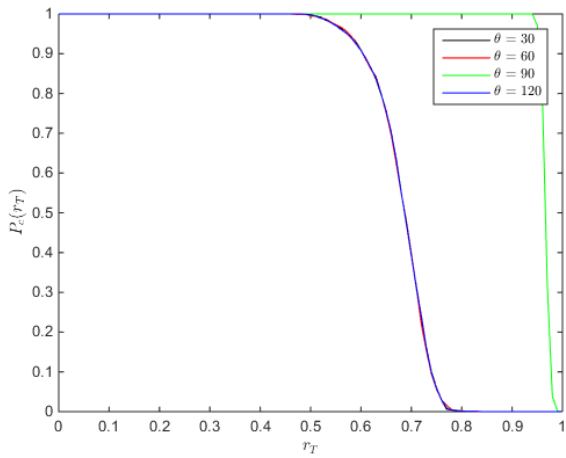

Figure 4. True positive versus matching ratio threshold for images rotated as much as 30, 60, 90, 120 degrees.

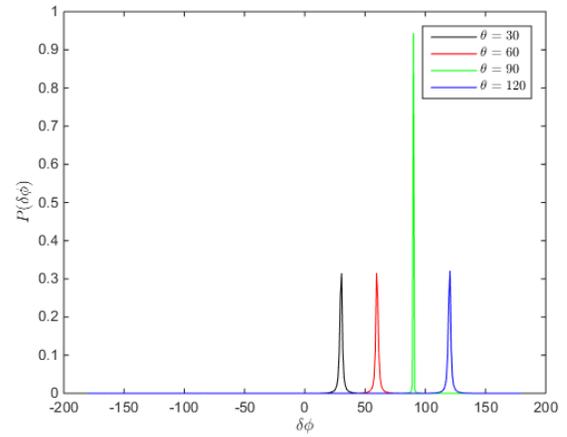

Figure 5. Probability of matched keypoint orientation difference for correct matches images rotated as much as 30, 60, 90, 120 degrees.

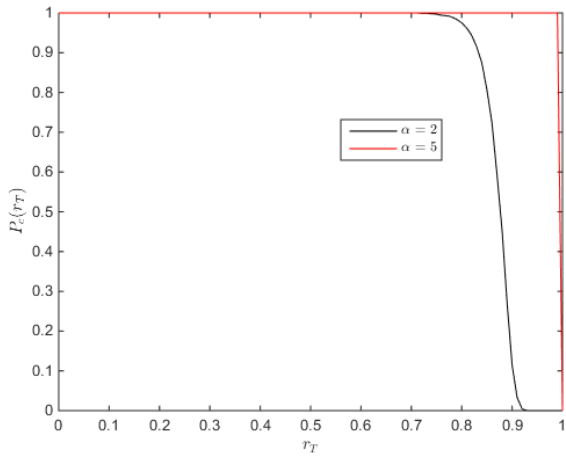

Figure 6. True positive versus matching ratio threshold for images with scaling factors $\alpha = 2$ and 5.

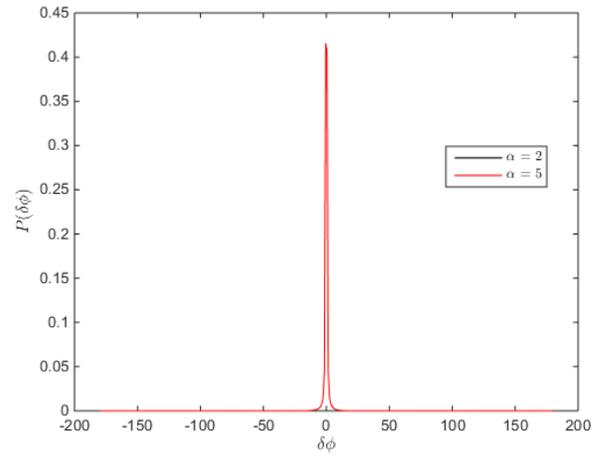

Figure 7. Probability of matched keypoint orientation difference for correct matches images with scaling factors $\alpha = 2$ and 5.

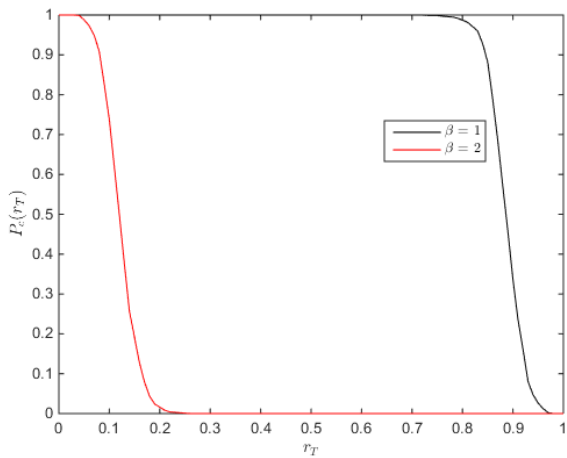

Figure 8. True positive versus matching ratio threshold for fish eye distorted images with distortion parameter $\beta = 1$ and 2.

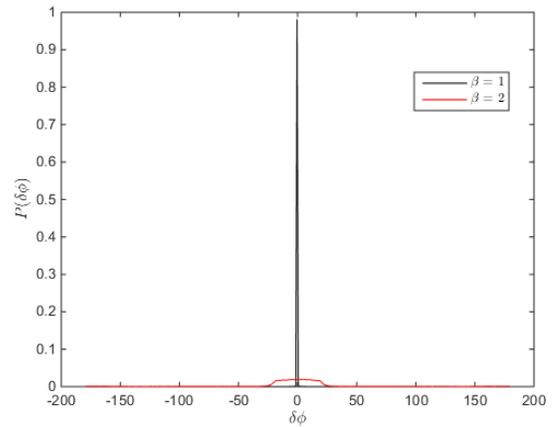

Figure 9. Probability of matched keypoint orientation difference for correctly matched fish eye distorted images with distortion parameter $\beta = 1$ and 2.

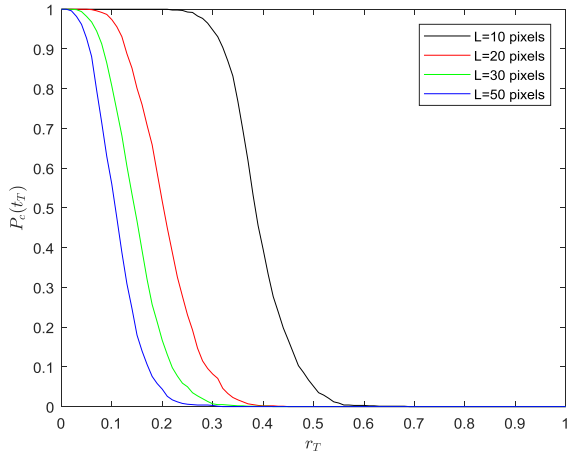

Figure 10. True positive versus matching ratio threshold for motion artifact as large as $L = 10, 20, 30,$ and 50 pixels.

### D. Fish Eye Distortion

Fish eye distortions are used for creating hemispherical panoramic images. There can be caused by lens of camera or manually created by using spherical distortions. Planetariums use the fish eye projection of night sky, flight simulations in order to create immersive environment for the trainee's uses the fish eye projection, some motion-picture formats also uses these projections. In meteorology fish eye lens are used to capture cloud formations.

For this experiment, 1000 images from Nister database are scaled with scaling factors $\beta = 1$ and 2, and true positive rate and $P(\delta\varphi)$ for each scenario are calculated. Figure 8 presents true positive rate versus matching rate threshold. From this figure, one can easily see that SIFT algorithm is very sensitive to fish eye distortion such that with $\beta = 2$, the true positive rate is very close to the false positive rate presented in figure 2 which results incorrect result. But from figure 9, one can see that $P(\delta\varphi)$ for fish eye distorted images is much different from the one for incorrect match as it is in figure 3. We can use this feature further to improve the performance of SIFT image matching against such distortion.

### E. Motion Artifact

Motion Artifact is a very common type of distortion and therefore, an image matching technique is required to be robust against it. For this experiment, motion artifact as large as $L = 10, 20, 30,$ and 50 pixels is applied to 1000 images from Nister database and true positive rate and $P(\delta\varphi)$ for each scenario are calculated. Figure 10 presents true positive rate versus matching rate threshold for images deformed with motion artifact. From this figure, one can easily see that SIFT algorithm is sensitive to motion distortion such that with $L = 50$, the true positive rate is very close to the false positive rate presented in figure 2 which results incorrect result. From figure 11, one can see that $P(\delta\varphi)$ for motion distorted images is even better than the one for fish eye distorted one, and there is a good potential for the optimization of SIFT image matching against such distortion.

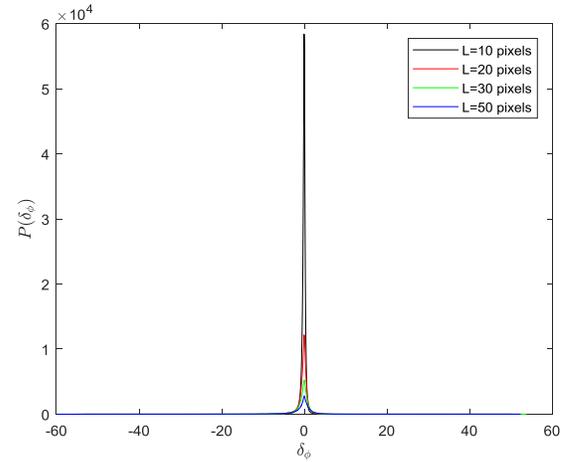

Figure 11. Probability of matched keypoint orientation difference for motion artifact as large as $L = 10, 20, 30,$ and 50

## IV. CONCLUSION

In this paper the performance of SIFT matching algorithm against various image distortion such as rotation, scaling, fish eye and motion distortion were evaluated and false and true positive rates for a large number of image pairs are calculated and presented. Also, the distribution of the keypoint orientation difference for correct and incorrect matches was presented. As the future work, the results obtained in this paper will be further used for optimization of SIFT matching accuracy.